\pdfoutput=1

\documentclass[a4paper,11pt,table,xcdraw]{article}
\usepackage{times,latexsym}
\usepackage{url}
\usepackage[T1]{fontenc}

\usepackage[acceptedWithA]{tacl2021v1}

\usepackage[]{hyperref}
\usepackage{xcolor}
\usepackage{times}
\usepackage{latexsym}
\usepackage[T1]{fontenc}
\usepackage[utf8]{inputenc}
\usepackage{microtype}
\usepackage{booktabs}
\usepackage{amssymb}
\usepackage{tabulary}
\usepackage{graphics}
\usepackage{todonotes}
\usepackage{linguex, tabularx}
\usepackage{multirow}
\usepackage{array}
\newcommand{\PreserveBackslash}[1]{\let\temp=\\#1\let\\=\temp}
\newcolumntype{R}[1]{>{\PreserveBackslash\raggedleft}p{#1}}

\newcommand{\nliex}[3]{
  \textbf{P:} #1\\
  \textbf{H:} #2  [E,N,C]: #3
}

\newcommand{\nliexs}[3]{
  \textbf{P:} #1  [E,N,C]: #3\\
  \textbf{H:} #2
}

\title{
  Investigating Reasons for Disagreement in
  Natural Language Inference}

\author{Nan-Jiang Jiang \\
  Department of Linguistics\\
  The Ohio State University \\
  USA\\
  \texttt{jiang.1879@osu.edu} \\\And
  Marie-Catherine de Marneffe \\
  Department of Linguistics / FNRS \\
  The Ohio State University / UCLouvain\\
  USA / Belgium\\
  \texttt{demarneffe.1@osu.edu} \\}

\begin{document}
\setlength{\Exlabelsep}{.5pt}
\setlength{\Exlabelwidth}{2em}
\setlength{\Extopsep}{0.4\baselineskip}

\maketitle

\begin{abstract}
We investigate how disagreement in natural language inference (NLI) annotation arises.
We developed a taxonomy of disagreement sources with 10 categories spanning 3
high-level classes.
We found that some disagreements are due to uncertainty in the sentence meaning, others to annotator biases and task artifacts, leading to
different interpretations of the label distribution.
We explore two modeling approaches for detecting items with potential disagreement: a 4-way
classification with a ``Complicated'' label in addition to
the three standard NLI labels, and a
multilabel classification approach.
We found that the multilabel classification is more expressive and gives better
recall of the possible interpretations in the data.

\end{abstract}

\section{Introduction}

Natural language inference (NLI) is the task of identifying whether a hypothesis sentence is inferred, contradicted, or neither, by a premise.
It is considered one of the most fundamental aspects of competent language understanding.
In natural language processing, the NLI task is widely used to evaluate
models’ semantic representations \citep[i.a.,][]{wang2018glue}, and to facilitate downstream tasks, e.g.\ in natural language generation (NLG).

Large NLI datasets have been built by collecting inference judgments for
premise-hypothesis pairs and aggregating the judgments by simple methods such as majority voting.
However, it has been pointed out that NLI items do not all have a single
ground truth and can exhibit systematic disagreement \citep[i.a.,][]{pavlick-kwiatkowski-2019-inherent,
  nie-etal-2020-learn}. This questions the assumption of having a
single ground truth for each item and the validity of measuring models' ability
to produce such ground truth.
For instance, in \ref{disagreement} from the MNLI dataset \cite{williams-etal-2018-broad}, 3 out of 5 annotators labeled the item as ``Entailment'' (the hypothesis is inferred from the premise), 0 labeled it as ``Neutral'' (the hypothesis cannot be inferred from the premise), and 2 as ``Contradiction'' (the hypothesis contradicts the premise).

{\small
  \ex.\label{disagreement}
  \nliex{the only problem is it's not large enough it only holds about i think they squeezed when Ryan struck out his five thousandth player they they squeezed about forty thousand people in there.}
  {It doesn't hold many people.}
  {[3,0,2]}

}

People have indeed different judgments on which number is required to count as \textit{holding many people}. The premise and hypothesis do not resolve explicitly what is being talked about, possibly a stadium. Does 40,000 count as \textit{many} for a stadium seating capacity? The premise states that \textit{it's not large enough} and uses the term \textit{squeezing}, leading some annotators to see the hypothesis \textit{it doesn't hold many people} as being inferred from the premise. On the other hand, 40,000 people in a specific location is a large number, and some annotators therefore judge the hypothesis as contradictory to the premise.
Such disagreement is not captured when taking only one of the three standard NLI labels as ground truth. Recent work \citep{zhang-etal-2021-learning-different,DBLP:journals/corr/abs-2104-08676} has thus explored approaches for building
NLI models that predict the entire annotation distribution, instead of the
majority vote category,
in an attempt to move away from assuming a single
ground truth per item.
However, little is understood about where the disagreement stems from, and whether modeling the distribution is the best way to handle disagreement in annotation.

To investigate these questions, we created a taxonomy of different types of disagreement consisting of 10 categories, falling
into 3 high-level classes based on the ``Triangle
of Reference'' by \citet{aroyo_truth_2015}.
We manually annotated a subset of MNLI with the 10 categories.
Our categorization shows that items leading to disagreement in annotation are highly heterogeneous. Moreover,
the interpretation of the NLI label distribution differs across items. We thus explored alternative approaches for modeling disagreement items: a 4-way
classification approach with an additional label (on top of the three NLI labels) capturing disagreement
items, and a multilabel classification approach of predicting one or more of
the three NLI labels.
We found that the two models behave somewhat differently, with the multilabel model offering more interpretable outputs, and thus being more expressive.
Our findings deepen our understanding of disagreement in a widely used NLI
benchmark and contribute to the growing literature on disagreement in
annotation.
We hope they highlight directions to reduce disagreement when collecting annotations and to design models to handle the disagreement that persists. The annotations, the guidelines and the code are available at \url{https://github.com/njjiang/NLI_disagreement_taxonomy}.

\section{Related work}
\label{sec:related}
Focusing on disagreement in annotation is not new: \citet{aroyo_truth_2015}
argued for embracing annotation disagreement, viewing it as signal, and not as
noise.
Even for tasks with supposedly a unique correct answer, such as part-of-speech tagging, there are items for which the right analysis is debatable \citep{plank-etal-2014-linguistically}: is
\textit{social} in \textit{social media} a noun or an adjective?
\citet{plank-etal-2014-learning} showed that incorporating such disagreement signal into the loss
functions of part-of-speech taggers improves performance.
Previous work noted that disagreement in annotation exists in many semantic tasks: anaphora resolution
\citep{poesio-artstein-2005-reliability,versley_vagueness_2008,poesio-etal-2019-crowdsourced},
coreference \citep{RECASENS20111138},
sentiment analysis \citep{kenyon-dean-etal-2018-sentiment},
word sense disambiguation
\citep{erk-mccarthy-2009-graded,10.1007/s10579-012-9188-x},
among others.

\paragraph{Sources for disagreement}
\citet{aroyo_truth_2015} introduced the ``Triangle of Reference'' framework to conceptualize the annotation process and explain annotation disagreement.
Annotation differences can stem from the sentences to be annotated, the labels, or the annotators. Indeed, annotators, who interpret the sentences, produce labels in a way that is defined by the annotation guidelines. Underspecification in each of these three components can result in disagreement in the annotations. Disagreement can arise from (1) uncertainty in the sentence meaning, (2) underspecification of the guidelines, (3) annotator behavior.
We use the Triangle of Reference to organize our taxonomy.%

\paragraph{Disagreement in NLI} \citet{demarneffe2012} and \citet{uma2021learning} showed that disagreement was systematic in the older NLI datasets. \citet{pavlick-kwiatkowski-2019-inherent} showed that real-valued NLI annotations are better modeled as coming from a mixture of Gaussians as opposed to a single Gaussian distribution.
\citet{nie-etal-2020-learn} collected categorical NLI annotations and found
disagreement to be widespread, corroborating \citet{pavlick-kwiatkowski-2019-inherent}'s findings.
\citet{kalouli-etal-2019-explaining} found that items involving entity/events coreference and ``loose definitions'' of inference (e.g.\ whether \textit{a hill covered by
  grass} is the same as \textit{the side of a mountain}) have lower
inter-annotator agreement.
However, there is not yet a systematic investigation of how disagreement in NLI arises.

\paragraph{Taxonomy in NLI}
There is a rich body of work on the taxonomy of reasoning types in NLI,
identifying the kinds of inferences exhibited in NLI datasets
\citep[i.a.,][]{sammons-etal-2010-ask,lobue-yates-2011-types,williams2020anlizing}.
Our work differs in that we focus on the phenomena that lead
to annotation disagreement, which are not necessarily reasoning types (e.g.\ our category Interrogative Hypothesis, \ref{ex:interrogative} in
Table~\ref{tab:category}).
Since we focus on disagreement, we do not categorize different ways of arriving at the same NLI label (e.g.\ different kinds of high agreement contradiction, as in \citet{de-marneffe-etal-2008-finding}).

\paragraph{Approaches to model disagreement}
\citet{pavlick-kwiatkowski-2019-inherent} argued that NLI disagreement information
should be propagated downstream. Current neural models should thus be evaluated
against the full label distribution.
Methods for approximating the full distribution have recently been developed for many tasks, using techniques for calibration and learning with soft-labels
\citep[i.a.,][]{https://doi.org/10.48550/arxiv.1702.08563,zhang-etal-2021-learning-different,fornaciari-etal-2021-beyond,DBLP:journals/corr/abs-2104-08676,10.3389/frai.2022.818451}.

However, simply because distributions are the most straightforward form of
disagreement information does not mean that they are the optimal representation for intrinsic evaluation or in downstream tasks.
Calibration techniques are successful at post-editing the classifier's softmax distribution \citep{pmlr-v70-guo17a},
but they convey spurious uncertainty for items that do not exhibit disagreement \citep{zhang-etal-2021-learning-different}.

Categorical decisions tend to be more interpretable and are
necessary in downstream tasks.
For example, NLI models are often used for automatic fact-checking \citep{thorne-etal-2018-fever,luken-etal-2018-qed}, where the categorical decision of whether a statement is disinformation determines whether it needs to be censored.
Therefore, we explore here different approaches for providing categorical information for disagreement.

For sentiment analysis, \citet{kenyon-dean-etal-2018-sentiment} used a classification approach with an
additional ``Complicated'' class to capture items with disagreement.
\citet{kenyon-dean-etal-2018-sentiment} had little success predicting that class with LSTM-based models (0.16 F1 for
Complicated), because it is heterogeneous and there is likely little
learning signals indicating complicatedness. \citet{zhang-de-marneffe-2021-identifying} approached the NLI 4-way
classification problem using the architecture of Artificial Annotator, an ensemble of multiple BERT models with different biases. They experimented on the NLI version of the CommitmentBank \citep{de_Marneffe_Simons_Tonhauser_2019}, and showed some success, obtaining 61.93\% F1 on the fourth class ``Disagreement'' using a vanilla-BERT baseline (standard fine-tuning BERT), and
66.5\% F1 on the ``Disagreement'' class using the Artificial Annotator architecture.
Here, we further test the 4-way classification approach for NLI.

In addition to its heterogeneity,
a ``Complicated'' or ``Disagreement'' class is not easily interpretable. We not only need to know whether there is disagreement, but also in what way: which labels do the annotators disagree over.
We therefore also take a multilabel classification approach
\citep[i.a.][]{10.1007/s10579-012-9188-x,oh-etal-2019-thomas,ferracane-etal-2021-answer},
predicting one or more of the three NLI labels.

There is another line of research aiming to model the judgments of individual annotators,
as opposed to the aggregated annotations representing the judgments of the population \citep{10.1145/3411764.3445423,10.1162/tacl_a_00449}.
However, these approaches require the annotators' identities for each annotation, which are often not released with the data.

\section{Disagreement taxonomy}
To investigate where disagreement stems from, we
conduct a qualitative analysis of parts of the MNLI dataset \citep{williams-etal-2018-broad}. We chose MNLI because it is diverse in genre and inference types, compared to
datasets based on image captions which only describe visual scenes (e.g.\ SICK \cite{marelli-etal-2014-sick}, SNLI \cite{bowman-etal-2015-large}).

\begin{figure}
  \centering
  \includegraphics[width=\linewidth]{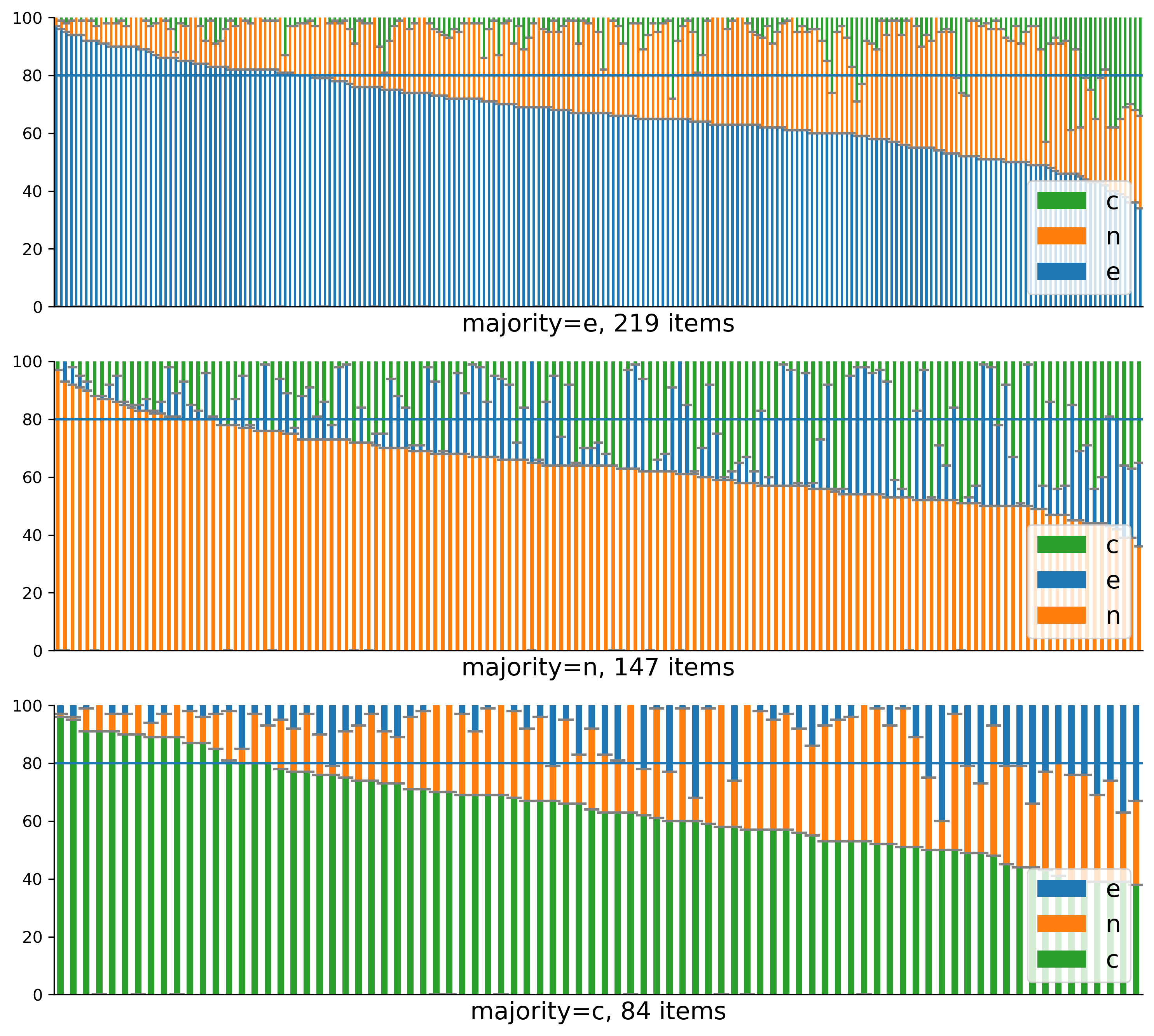}
  \caption{ChaosNLI annotations of the 450 items we sampled. Each column of stacked bars represents an item's annotations -- the number of votes for each label with
    top-down ordering of the labels.
    The horizontal lines indicate 80 votes.
    }
  \label{fig:tax_stack}
\end{figure}
\subsection{Data to analyze}
The original MNLI dev sets (match and mismatch sets, differing in
genres)\footnote{The match dev set is of the same genres as the training set
  (e.g.\ fiction, government websites), while the
  mismatch dev set comes from other genres than the training set (e.g.\
  face-to-face conversations, letters).} contain 5 annotations per item.
The MNLI dev matched set contains 1,599 items for which exactly 3 annotators (out
of the 5) agreed on the label. This subset was reannotated by
\citet{nie-etal-2020-learn} with 100 annotations per item, called the ChaosNLI
dataset. We randomly sampled 450 items from ChaosNLI. Figure~\ref{fig:tax_stack} shows the annotations, with items organized by which label was the most frequent. While some items can be seen as having a unique ground truth label (depending on how many annotators agreeing on the same label are needed for that -- here we take 80\%, following \citet{jiang-de-marneffe-2019-evaluating}), other items clearly lead to differing annotations.

We also sampled 60 items from the MNLI dev matched set
in which at most 2 out of the 5
annotators agreed on the label, and there are thus no majority labels.
These items, coded with label ``\texttt{-}'' in \citet{williams-etal-2018-broad}'s release, are customarily discarded in evaluating NLI models.

\newcounter{myexample}
\renewcommand{\themyexample}{[\arabic{myexample}]}
\newcommand{\example}{\leavevmode\refstepcounter{myexample} \themyexample}

\begin{table*}
  \small
\resizebox{\linewidth}{!}{
\begin{tabular}[t]{@{}lp{3.6cm}p{9cm}p{7cm}p{1cm}p{1.1cm}}
  \toprule
                                            &  & Premise                                                                                                                                                                                                                                                               & Hypothesis                                                                            & MNLI [E,N,C] & ChaosNLI [E,N,C] \\
  \midrule
  \multicolumn{3}{l}{\textbf{Uncertainty in Sentence Meaning}} \\
  \midrule

 \example \label{ex:lex_pair}&Lexical & Technological advances generally come in waves that crest and eventually subside. & Advances in electronics come in waves. & [3,1,1] & [82,17,1]\\
\example \label{ex:impl}&Implicature                       & Today it is possible to buy cheap papyrus printed with gaudy Egyptian scenes in almost every souvenir shop in the country, but some of the most authentic are sold at The Pharaonic Village in Cairo where the papyrus is grown, processed, and hand-painted on site. & The Pharaonic Village in Cairo is the only place where one can buy authentic papyrus. & [0,2,3]          & [2,39,41]          \\
  \example \label{ex:presupposition}&Presupposition & What changed?	& Nothing changed. &[0,2,3] & [4,76,20] \\
\example \label{ex:enrich}&Probabilistic Enrichment                    & It's absurd but I can't help it. Sir James nodded again.                                                                                                                                                                                                              & Sir James thinks it's absurd.                                                         & [3,2,0]          & [63,35,2]         \\
\example\label{ex:imperfection}&Imperfection & profit rather	& Our profit has not been good. & [0,3,2] & [3,90,7] \\
  \midrule
  \multicolumn{3}{l}{\textbf{Underspecification in Guidelines}} \\
  \midrule
  \example\label{ex:coref}&Coreference
  & This was built 15 years earlier by Jahangir's wife, Nur Jahan, for her father, who served as Mughal Prime Minister.  & Nur Jahan's husband Jahangir served as Mughal Prime Minister.& [2,0,3]& [24,45,31]\\
\example \label{ex:temporal}&Temporal Reference                            & However, co-requesters cannot approve additional co-requesters or restrict the timing of the release of the product after it is issued.                                                                                     & They cannot restrict timing of the release of the product.                                           & [3,2,0]        & [90,8,2]                   \\
  \example\label{ex:interrogative}&Interrogative Hypothesis &
  How did you get it?" A chair was overturned. & "How did you get your hands on this object?" & [3,2,0] & [45,52,3] \\
  \midrule
  \multicolumn{3}{l}{\textbf{Annotator Behavior}} \\
  \midrule
  \example\label{ex:underspec}&
  Accommodating Minimally Added Content & Indeed, 58 percent of Columbia/HCA's beds lie empty, compared with 35 percent of nonprofit beds.                                                                                                                                                                      & 58\% of Columbia/HCA's beds are empty, said the report.                                & [3,2,0]          & [97,3,0]          \\
\example\label{ex:np_shuffle}&  High Overlap   & Yet, in the mouths of the white townsfolk of Salisbury, N.C., it sounds convincing.	& White townsfolk in Salisbury, N.C. think it sounds convincing.
                 & [3,2,0]          & [68,27,5]         \\
  \bottomrule
\end{tabular}
}
\caption{Categories of potential sources of disagreement, with
  examples. The last two columns give the number of annotations for each NLI label ``Entailment'' (E), ``Neutral'' (N), and ``Contradiction'' (C), in MNLI and ChaosNLI.}
\label{tab:category}
\end{table*}

\subsection{Disagreement categories}
\label{section:taxonomy}

Our taxonomy of potential disagreement sources consists of 10 categories, shown in Table~\ref{tab:category}.
The categories are organized into three high-level classes, corresponding to the three components of the annotation process in the ``Triangle
of Reference'': (1) uncertainty in the sentence meaning, (2) underspecification of the guidelines, (3) annotator behavior.

\subsubsection{Uncertainty in sentence meaning}
Some textual phenomena leading to
disagreement can be local to \textbf{Lexical} items, where the truth
of the hypothesis depends on the meaning of a specific lexical item.
That lexical item can have multiple meanings, or its meaning requires certain
parameters that remain underspecified in the sentence at hand, as we
saw with \textit{many} in \ref{disagreement}.\footnote{
  It is challenging to distinguish between multiple senses or implicit parameters. For instance, in the pair
  \textit{P: Then he sobered.}	- \textit{H: He was drunk.}, whether H can be inferred from P depends on the word \textit{sober}: one could be sober from alcohol or from other drugs. Are these two meanings of the word or is the substance an implicit parameter?}
Disagreement can come from a pair of lexical items, where the lexical relationship between the items (e.g.\ hypernymy, synonymy) is loose, as in \ref{ex:lex_pair} in Table~\ref{tab:category}: do people infer \textit{advances in electronics} from \textit{technological advances}?

Other cases involve the holistic meaning of the sentences and interpreting them in different contexts. In some cases, the hypothesis is an \textbf{Implicature} of the premise, as in \ref{ex:impl}. By definition, an implicature can be cancelled \citep{grice1975logic}, which leads to a potential for differences in the annotations. %
Here, \textit{some of the most authentic papyrus (are sold in The Pharaonic Village)} gives rise to the scalar implicature \textit{but not all of the most authentic papyrus}, making the hypothesis false since it asserts that authentic papyrus are only sold in The Pharaonic Village. However if the implicature is cancelled, \textit{some} can also be interpreted as \textit{all} (e.g.\ \textit{Some students came. In fact, all came.})

The hypothesis can target what is being \textbf{presupposed} by the premise.
Wh-questions, for instance, presuppose that the entity the
question bears on exists. The question \textit{What changed?} in \ref{ex:presupposition} presupposes that something changed, hence the answer \textit{Nothing changed} can be viewed as contradictory. However, the premise can also be viewed as not giving enough information to judge the truth of the hypothesis, which would lead to a Neutral label.

 \textbf{Probabilistic Enrichment} items involve making probabilistic inferences from the premise: the inferred content is likely, but not definitely, true in some contexts. In \ref{ex:enrich}, there is some likelihood that nodding to the speaker's assertion means that one agrees with it. If annotators make that inference, they see the hypothesis as Entailment. But, since the premise is not explicitly stating the hypothesis, a Neutral label is also warranted.

Some premises/hypotheses contain typos or are fragments, making it hard to grasp their exact meaning (as in \ref{ex:imperfection}). We call these cases \textbf{Imperfection}, following \citet{williams2020anlizing}.

\subsubsection{Underspecification in the guidelines}
Some disagreements stem from the loose definition of the NLI task.
Assuming \textbf{coreference} between the premise and the hypothesis has been
noted as an important aspect of the NLI task \citep{mirkin-etal-2010-assessing}
and necessary for obtaining high agreement in annotation
\citep{de-marneffe-etal-2008-finding,bowman-etal-2015-large,kalouli-etal-2019-explaining}. %
In \ref{ex:coref}, the hypothesis is a contradiction if we assume that \textit{Mughal Prime Minister} is the same person in both the premise and the hypothesis. However, it could be the case that Nur Jahan's father
and husband both served as Mughal Prime Minister but in different terms, making it Neutral.

While the NLI task assumes coreference between entities and events mentioned in the premise and hypothesis, which entity/event to take into consideration is not always clear. For example, in \ref{ex:unfortunately}, the premise can be taken to talk about ``desegregation being undone in Charlotte by magnet schools'', in which case the hypothesis is inferred.

\ex.\label{ex:unfortunately}
\nliex{
Unfortunately, the magnet schools began the undoing of desegregation in Charlotte.
}
{Desegregation was becoming disbanded in Charlotte thanks to the magnet schools.}
{[81, 6, 13]}

The premise can also be taken to focus on the fact that ``the desegregation being undone in Charlotte by magnet schools is unfortunate''. In other words, two different ``Questions Under Discussion'' \cite{roberts_information_2012} can be posited for the premise. Under that second interpretation, the hypothesis (in which the undoing of desegregation is positive, given the word \textit{thanks}) contradicts the premise, where the desegregation undoing is unfortunate.

The truth of the hypothesis can also depend on the time at which the hypothesis
is evaluated (\textbf{Temporal Reference}), but the NLI annotation guidelines do not specify how to handle such cases.
There are two contextually-salient temporal referents in \ref{ex:temporal},
before or after the product release is issued.
If the hypothesis refers to the time after the release is issued, it is true.
From the perspective of before the release is issued, it is unclear whether the co-requesters can restrict timing or not.

Unlike assertions, questions do not have truth values
\citep{groenendijk1984studies,roberts_information_2012}.
It is therefore theoretically ill-defined to
ask whether an \textbf{interrogative hypothesis} is true or not given the premise (which is the question asked in \citet{nie-etal-2020-learn}'s annotation interface to build ChaosNLI). However, most of the interrogative hypotheses have interrogative premises (81.8\% in MNLI dev sets; all in our subset). \citet{groenendijk1984studies} define the notion of entailment between questions: an interrogative \textit{q1} entails
another \textit{q2} iff every proposition that answers \textit{q1} answers \textit{q2} as well. Some annotators seem to latch onto this definition, as in \Next.

{
  \small
  \ex. \nliex{yeah but uh do you have small kids}{Do you have any children?}{[65,33,2]}

}

Still, there is no definition distinguishing neutral from contradictory pairs of questions.\footnote{The issue does not necessarily arise from interrogative premises, since the hypothesis may target the presupposition of the question, as in \ref{ex:presupposition}.}
Annotators, perhaps to assign some meaning to the Neutral/Contradiction distinction, give
judgments that seem to involve applying surface-level features for declarative sentences, choosing Neutral/Contradiction if the sentences involve substitution of unrelated words, as in \Next.

{
  \ex. \nliexs{Where is art?}	{What is the place of virtue?}{[1,59,40]}

}

\subsubsection{Annotator behavior}
By definition, disagreement arises when a proportion of annotators behave one way and another proportion another way. We identified two patterns of ``systematic behavior'' (while it is hard to say for certain what annotators have in mind, the patterns seem robust). When the hypothesis adds content that provides minimal information compared to the premise, but is otherwise entailed, annotators are more likely to judge it as Entailment, thus
ignoring/\textbf{accommodating minimally added content}.
For instance, the hypothesis in \ref{ex:underspec} adds the information source (\textit{said in the report}) which is not mentioned in the premise. From a strict semantic evaluation, the hypothesis is thus not inferred from the premise. Nonetheless most people are happy to infer it.  Such added contents are often not at-issue, i.e.\ not the main point of utterance
\cite{Potts05BOOK,simons2010projects}, appearing as
modifiers \citep{mcnally_2016}, or parentheticals, making it easier for people to ignore if not paying enough attention or not being tuned to such differences.\footnote{Items belonging to other categories may also exhibit such pattern, such as \ref{ex:coref} for which 24/100 annotators chose Entailment.
Note that annotators for ChaosNLI were carefully vetted and passed multiple screening and training rounds.}

These biases are potentially problematic for applications in NLG that
use the NLI labels for evaluating paraphrases (modeled as bi-directional
entailment, \citet{ws-2007-acl-pascal}), dialog coherence
\cite{dziri-etal-2019-evaluating-coherence}, semantic accuracy
\cite{dusek-kasner-2020-evaluating}, or use NLI as a pretraining task for
learned metrics \cite{sellam-etal-2020-bleurt}.
For instance, it would not be semantically accurate for a generated summary to hallucinate and include extraneous, even if not at-issue content, such as \textit{said the report} in \ref{ex:underspec}, if not already given in the source text.

When the hypothesis has \textbf{high lexical overlap} with the premise (e.g.\ involve the same
noun phrases), annotators tend to judge it as Entailment even if it is not strictly inferred from the premise. In \ref{ex:np_shuffle}, the hypothesis claims that the white townsfolk
thinks it sounds convincing, whereas the premise only states that the white townsfolk makes it sound convincing (and does not mention whose opinion it is). \citet{mccoy-etal-2019-right} pointed out that
items in MNLI with high lexical overlap between the premise and the hypothesis often have
the Entailment label, and that NLI models learn such shallow heuristics, ending up to
incorrectly predict Entailment for items with high overlap.
\citet{mccoy-etal-2019-right}'s finding might partially be attributed to such
annotator behavior.

\subsection{Taxonomy development and annotation}
The taxonomy was developed by a single annotator, starting by examining lowest and highest agreement examples in ChaosNLI to identify linguistic phenomena that are potential sources of disagreement in the NLI annotations.
Some categories were merged because the distinction between them seem murky (for instance, the distinction of multiple senses vs.\ implicit argument in the Lexical category). Event coreference often requires entity coreference and the distinction between both is not clear-cut. For the two sentences \textit{vendors crammed the streets with shrine offerings} and \textit{vendors are lining the streets with torches and fires} to refer to the same event, we need to assume that they talk about the same set of vendors. We thus only have one Coreference category.

There were two rounds of annotations. In Round 1, one annotator
annotated 400 items from ChaosNLI and iteratively refined the taxonomy, while
writing annotation guidelines. Another annotator was then trained. In Round 2, both annotators annotated 50 additional items from ChaosNLI and 60
items from MNLI where only 2 out of the 5 original annotations agreed. These 110 items serve to check that the taxonomy does not ``overfit'' the 400-item sample used while developing it.

\paragraph{Multi-category annotations}
More than one reason for disagreement may apply. We therefore adopt a multi-category annotation scheme: each item can have multiple categories. For example in \Next, both Implicature and Temporal Reference contribute to disagreement. The premise does not suggest that the park changed name, while the hypothesis does so with the implicature triggered by \textit{used to}. Therefore, if we evaluate the truth of the hypothesis now, there can be disagreement between Neutral and Contradiction. If we evaluate the truth of the hypothesis in or before 1935, the hypothesis is entailed because the park was named after Corbett at some point. Also, given that the implicature is triggered by a specific lexical item (in contrast to non-conventional conversational implicatures), the category Lexical applies too.

{\small
\ex. \nliex{The park was established in 1935 and was given Corbett's name after India became independent.}{The park used to be named after Corbett.}{[36, 34, 30]}

}

\paragraph{Inter-annotator agreement}
Since the annotation requires the understanding of various
linguistic phenomena, only expert annotation is possible. The two annotators have graduate linguistic training.
The Krippendorff's $\alpha$ with MASI distance \citep{passonneau-2006-measuring} is 0.69. For the items annotated by both annotators, we then aggregated the two sets of annotations by taking their intersection. 
This resulted in 24 instances of categories deleted for annotator 1 (in 23 items) and 16 (in 16 items) for annotator 2. There were only 4 items (out of 110) with an empty intersection, which we reconciled.

\paragraph{Distribution of categories}
Table~\ref{tab:cat_freq} shows the frequency of each combination of categories
for the two rounds of annotations.
Probabilistic Enrichment and Lexical are the two most frequent categories, because they are broad categories by definition and not tied to a close set of lexical items.
It also shows that the Round 2 annotations have roughly the same
distribution, although some combinations are rarer/did not appear in Round 1. No
items have been encountered in Round 2 that needed creation of a novel disagreement category.

There are 11 items in the Round 1 sample for which none of the annotators could identify a source of disagreement.
These items exhibit characteristics of clear, easily-identifiable cases:
paraphrase or containment relations, for Entailment \ref{ex:clear_e}; antonym or negation, for
Contradiction \ref{ex:clear_c}; statements containing information that is not given by nor can be inferred from the premise, for Neutral \ref{ex:clear_n}.
They also involve high agreement in the ChaosNLI annotations (average number of majority votes between 65 and 95, with a mean of 84.1).

{
\small
  \ex.\label{ex:clear_e}
  \nliex{
  well so okay you need to get married and have kids and then when they're big
  enough you can have them go do the yard and you can do what you want to do}
{When your kids grow up you have have them do the yardwork.}{[86,10,4]}

  \ex.\label{ex:clear_c}
  \nliex{
    uh-huh you can't do that in a skirt  poor thing
  }{You can do anything in a skirt.}{[3, 23, 74]}

  \ex.\label{ex:clear_n}
  \nliex{She had the pathetic aggression of a wife or mother--to Bunt there was no difference.}{Bunt was raised motherless in an orphanage.}{[0,88,12]}

}

\begin{table}[t]
\centering
\resizebox{\linewidth}{!}{
  \begin{tabular}{lrrrr}
\toprule
 & \multicolumn{2}{c}{Round 1} & \multicolumn{2}{c}{Round 2} \\
 &    \# &     \%              &      \# &     \% \\
\midrule
Probabilistic              &  113 &  29.05 &   31 &  28.18 \\
Lexical                    &   65 &  16.71 &   32 &  29.09 \\
Coreference                &   48 &  12.34 &   14 &  12.73 \\
Accommodating              &   46 &  11.83 &    5 &   4.55 \\
Imperfection               &   18 &   4.63 &    4 &   3.64 \\
Lexical, Probabilistic     &   17 &   4.37 &    1 &   0.91 \\
Interrogative.             &   15 &   3.86 &    5 &   4.55 \\
Implicature, Lexical       &   11 &   2.83 &    0 &   0.00 \\
Implicature                &    8 &   2.06 &    0 &   0.00 \\
Coreference, Probabilistic &    8 &   2.06 &    4 &   3.64 \\
High Overlap               &    7 &   1.80 &    1 &   0.91 \\
Presupposition             &    6 &   1.54 &    3 &   2.73 \\
Temporal                   &    5 &   1.29 &    2 &   1.82 \\
Coreference, Lexical       &    5 &   1.29 &    0 &   0.00 \\
Lexical, Presupposition    &    3 &   0.77 &    0 &   0.00 \\
Implicature, Probabilistic &    2 &   0.51 &    1 &   0.91 \\
Coreference, Imperfection  &    1 &   0.26 &    2 &   1.82 \\
Coreference, Temporal      &    1 &   0.26 &    1 &   0.91 \\
Lexical, Temporal          &    1 &   0.26 &    1 &   0.91 \\
Probabilistic, Temporal    &    1 &   0.26 &    1 &   0.91 \\
    \midrule
Sub-Total                      &  381 &  97.98 &  108 &  98.21 \\
    \midrule
    \multicolumn{5}{l}{\textbf{8 combinations occurred once only in Round 1}} \\
    \multicolumn{5}{p{1.22\linewidth}}{
    Accommodating, Probabilistic | Presupposition, Probabilistic | Lexical, Presupposition, Probabilistic | Accommodating, Lexical, Probabilistic | Imperfection, Lexical | Accommodating, Lexical | Coreference, Implicature | Implicature, Temporal
    } \\
    \midrule
    \multicolumn{5}{l}{\textbf{2 combinations occurred once only in Round 2}} \\
    \multicolumn{5}{p{1.235\linewidth}}{Presupposition, Temporal | High Overlap, Lexical, Probabilistic} \\
\bottomrule
\end{tabular}
}
\caption{Frequency and percentage of each combination of categories in the
  taxonomy, in the two annotation rounds. %
  }
\label{tab:cat_freq}
\end{table}

\subsection{Findings and discussion}\label{findings}

Through the construction of the taxonomy, we found that disagreement arises from
many reasons. The NLI annotations do not always show the full picture in terms
of the range and nature of the meaning the sentences carry, because (1) even if an item has multiple possible interpretations, the annotators may converge on
one of them, (2) there are at least two interpretations of the label distribution, arising out of a single probabilistic
inference, or multiple categorical inferences.

\paragraph{Annotators converge on one interpretation}
NLI annotations for items exhibiting some of the factors that contribute to disagreement may actually show high agreement.
Indeed, even when an item lends itself to uncertainty or multiple interpretations, a high proportion of annotators may converge to the same interpretation. For instance,
in \ref{ex:lex_pair} (Table~\ref{tab:category}), 82 annotators (out of 100) take \textit{technological advancement} to entail \textit{advancement in electronics},
even though there are other kinds of technological advancement that are not electronics.
In \ref{ex:imperfection}, 90 annotators latch onto the fact that the hypothesis
seem totally unrelated to the premise, agreeing on the Neutral label.

\begin{table}[t!]
\centering
\resizebox{\linewidth}{!}{
\begin{tabular}{lrrr}
\toprule
 &  Converge \% &   Total \# &     Mean (std) \\
 & & & majority vote \\
\midrule
Lexical &     17.74 &  124 &  66.02 (14.15) \\
Implicature &     12.50 &   24 &  63.96 (14.21) \\
Presupposition &      0.00 &   12 &  57.92 (13.82) \\
Probabilistic Enrichment &     13.33 &  165 &  64.56 (12.06) \\
Imperfection &     22.73 &   22 &  67.18 (14.71) \\
Coreference &     14.67 &   75 &  66.17 (14.39) \\
Temporal Reference &     25.00 &   12 &   62.0 (19.33) \\
Interrogative Hypothesis &     20.00 &   15 &  63.13 (14.32) \\
Accommodating &     25.49 &   51 &  67.76 (16.48) \\
High Overlap &      0.00 &    8 &   65.12  (4.75) \\
\bottomrule
\end{tabular}
}
\caption{For each disagreement category, the percentage of items exhibiting convergence (at least 80/100 annotators agreed on the same NLI label), the total number of items in the category, and the mean/standard deviation of the majority vote count.}
\label{tab:converge}
\end{table}

Table~\ref{tab:converge} shows the percentage of items in each
taxonomy category for which at least 80 (out of 100) annotators agreed on the same NLI label (which we will refer to as ``convergence'').
Interestingly, ``Accommodating minimally added content'' has the largest amount of convergence (25.5\%) and the highest mean majority vote (67.8).
The majority voted labels are Entailment (accommodating the content) or Neutral
(considering that the content is not given by the premise).
Whether accommodation takes place depends on the extent to which the added content is not-at-issue and on the content itself. In \ref{ex:underspec} (Table~\ref{tab:category}), 97 annotators accommodated
the extra content (\textit{said the report}) in the hypothesis.
In \ref{ex:muc_no_accommodate}, however, the hypothesis also introduces new content
\textit{all year round}, but only 7 annotators
accommodate it.
In \ref{ex:muc_half_accommodate}, 32 annotators accommodate the added content
\textit{American}, thus more than in \ref{ex:muc_no_accommodate} but less than in
\ref{ex:underspec}.

{
\small
\ex.\label{ex:muc_no_accommodate}
\nliex{
The equipment you need for windsurfing can be hired from the beaches at Tel Aviv
(marina), Netanya, Haifa, Tiberias, and Eilat.}
{Windsurfing equipment is available for hire in Tel Aviv all year round.}
{[7, 93, 0]}

\ex.\label{ex:muc_half_accommodate}
\nliex{
  And here, current history adds a major point.
}
{Current American history adds a major point.}
{[32, 67, 1]}

}

The difference could be due to the fact that
\textit{American} modifies the subject, which makes it less at-issue than
\textit{all year round} modifying the entire matrix clause.
In \ref{ex:underspec}, \textit{said the report} appears in a parenthetical at
the end of the sentence, which is even less at-issue than modifiers. Identifying such gradience in disagreement is a very difficult task: simply identifying whether the hypothesis adds content is not enough, knowledge about the role of information structure seems necessary too.

\paragraph{Two interpretations of NLI label distributions}
It should now be clear that, by modeling majority
vote, we are missing out on the full complexity of language understanding.
Some argue that textual inference is probabilistic in nature \cite{glickman-dagan-2005-probabilistic}.
Therefore, probabilistic inferences give rise to disagreement in categorical labels. %
\citep[i.a.][]{zhang-etal-2021-learning-different,DBLP:journals/corr/abs-2104-08676}.
Here, we found that disagreement in the categorical labels arise in at least two
ways:
(1) a single probabilistic
inference, or (2) multiple potentially categorical inferences,
which is often the case when there are multiple possible specifications of the contextual
factors (e.g.\ coreference, temporal reference, implicit arguments of some lexical items).
They also differ in the kinds of uncertainty they exhibit. One is uncertainty in
the state of the world. One is in how to interpret the sentences.

This distinction gives different interpretations of the aggregated label distribution.
In \ref{ex:enrich} (Table~\ref{tab:category}), each annotator may have an underlying probabilistic judgment of how likely it is that \textit{Sir James thinks it's absurd}, which is then reflected in the aggregated probability distribution.
The probability associated with the Entailment label can be taken as
the probabilistic belief \citep{kyburg1968bets} of an individual annotator for the truth of the hypothesis.

On the other hand, \ref{ex:temporal} involves categorical and probabilistic inferences. Whether the hypothesis is entailed depends on whether it is evaluated
before or after the product release is issued.
If after, readers have a \textbf{categorical}
judgment that the hypothesis is entailed.
If before, readers have a \textbf{probabilistic} judgment, leading
to the uncertainty between Neutral and Contradiction.
Therefore, unlike \ref{ex:enrich}, the probability associated with the
Entailment label does not represent the judgment of an individual.

We could design experiments to collect empirical evidence for this distinction, such as collecting multilabel or sliding bar annotations, or free-text
explanations to gain direct evidence of whether annotators have categorical/probabilistic judgments. Pursuing this line of research is left for
future work.

\paragraph{Artificial task setup}
One of the reasons for the occurrence of disagreement may be the somewhat artificial setup of the NLI task.
The premise and hypothesis are interpreted in isolation with no surrounding discourse. However, discourse context is needed for resolving many of the uncertainty in meaning pointed out here (e.g.\ coreference, temporal reference, and implicit arguments of lexical items).
Investigating whether incorporating context into NLI annotations improves agreement is left for future work. 

\section{Modeling experiments}

Now that we understand better how disagreement arises, we explore how to build
models that provide disagreement information.
As discussed in Section~\ref{sec:related}, a distribution gives the most
fine-grained information but can be misleading to interpret, while categorical
information is often needed in downstream applications.
Therefore, we experiment with models that provide two kinds of categorical
information for disagreement: an additional ``Complicated'' class for
labeling low agreement items (Section~\ref{sec:4way}), and a multilabel classification approach, where each item is
associated with one or more of the three standard NLI labels (Section~\ref{sec:multilabel_classification}).
As baseline, we take the MixUp approach in
\citet{zhang-etal-2021-learning-different}, which predicts a distribution over
the three labels, and uses a threshold to obtain multilabels/4-way labels.

These models can be useful in an annotation pipeline. One needs to collect multiple judgments for each item to cover the range of possible interpretations, but doing so may be prohibitively expensive at a large scale.
The annotation budget could thus be prioritized by collecting annotations for items
with potential for disagreement, as predicted by the model. Therefore, our goal is not necessarily to maximize accuracy. A model that can recall the possible interpretations is preferred to a model that misses them.

\subsection{Training data}
\label{sec:training_data}
We saw that there is gradience in disagreement, but we start with clearly
delineated data and only take items for which there is distinct (dis)agreement.
We first focus on items from ChaosNLI since they have 100 annotations each, giving a
clearer signal for (dis)agreement, discarding items where the majority vote is between 60 and 80 (given that it is unclear whether this counts a high or low agreement). However, this gives a highly class-imbalanced set in both schemes, as shown in
the line for ``Chaos'' in Table~\ref{tab:class_size}, with
less items in E/N/C than in the other classes.\footnote{Both the baseline and
  our models perform poorly when trained with the imbalanced set.}
Therefore, we augment the set with data from the original MNLI dev sets (where items have 5 annotations).
We use the following criteria to relabel the data with the 4-way scheme (E, N, C, and Complicated) and the multilabel scheme:
\begin{itemize}
\setlength\itemsep{-0.1cm}
    \item[-] Items receive a single E, N, or C label (in the 4-way and multilabel schemes) if the majority vote label has more than 80 votes (out of 100 annotations) for the ChaosNLI items or if all 5 annotations agree for the MNLI items.
    \item[-] ChaosNLI items are labeled as Complicated or as having multiple labels if the majority has less than 60 votes. For multilabel, a label is present if it has at least 20 votes (complement of 80 used for the single label items). MNLI items are labeled as Complicated if two labels have at least 2 votes (e.g.\ [3,2,0] or [2,2,1]). For multilabel, a label is present if it has at least 2 votes.
\end{itemize}

We downsampled\footnote{We also experimented without downsampling majority vote:
  it worsened the performance on identifying the Complicated class or items with
  multiple labels.}
MNLI items with one of the E/N/C labels to get a class-balanced set in
the multilabel scheme. The resulting sizes are shown in Table~\ref{tab:class_size}, line ``Chaos+Orig''. We split the ``Chaos+Orig'' set into train/dev/test with sizes 2710/816/1956
respectively, stratified by labels.

\begin{table}

  \resizebox{\linewidth}{!}{
\begin{tabular}{lrrr|rrrr}
  \toprule
Dataset & E    & N    & C    & EN   & NC  & EC  & ENC  \\
  \midrule
\multirow{2}{*}{Chaos}      & \multirow{2}{*}{195}  & \multirow{2}{*}{57}   & \multirow{2}{*}{37}   & 291       & 205    & 32     & 76     \\
                            &                       &                       &                       & \multicolumn{4}{c}{604 Complicated}  \\
\multirow{2}{*}{Chaos+Orig} & \multirow{2}{*}{1117} & \multirow{2}{*}{1117} & \multirow{2}{*}{1117} & 1117      & 775    & 163    & 76     \\
                            &                       &                       &                       & \multicolumn{4}{c}{2131 Complicated} \\
                                                                                                      \bottomrule
\end{tabular}
}
\caption{Number of items for each 4-way label and each combination of
  multilabel in each dataset.
  The number of ``Complicated'' items is the sum of the number of items with more than
  one label in the multilabel setup.
}
\label{tab:class_size}
\end{table}

\subsection{Baseline}
We use \citet{zhang-etal-2021-learning-different}'s MixUp model as baseline for both the 4-way and multilabel schemes. The MixUp model has the same architecture as fine-tuning RoBERTa for
classification. 
During training, each training example is a linear interpolation of two randomly chosen training items, for both the input encodings and the soft-labels (the annotation distributions over E/N/C). We used \citeauthor{zhang-etal-2021-learning-different}'s hyperparameters, with a learning rate of 1e-6 and an early stopping patience of 5 epochs. The model is trained with the data split described above by optimizing
KL-divergence with soft-labels.

To evaluate, we convert each predicted distribution to a multilabel, taking any label assigned a probability of at least 0.2 to be present (same threshold we used
for the data).
The multilabel is then converted to a 4-way label: Complicated if more than one label is present; E, N, or C if it is the only label. Comparing the results from the MixUp model with the ones from our approach will tell whether optimizing for distributions (as done by the MixUp model) gives better predictions than training with categorical labels (as done by our approach), when evaluating with categorical labels.

\subsection{4-way classification}
\label{sec:4way}
We fine-tuned RoBERTa \cite{liu2019roberta} on the train/dev
set using the standard methods for classification.
We used the initial learning rate of 1e-5, with learning rate decay by 0.8 times if dev F1 does not improve for two epochs.
We trained for up to 30 epochs, with early stopping used if dev F1 does not improve for 10 epochs.
We used \texttt{jiant v1} \cite{wang2019jiant} for our experiments.

\begin{table}
\centering
\resizebox{\linewidth}{!}{
\begin{tabular}{lrr|rrrr}
\toprule
 & \multicolumn{2}{c|}{Chaos+Orig} & \multicolumn{2}{c}{Chaos} & \multicolumn{2}{c}{Orig} \\
 & MixUp & Our & MixUp & Our & MixUp & Our \\
\midrule
Accuracy & {\cellcolor[HTML]{72B2D8}} \color[HTML]{000000} 63.50 & {\cellcolor[HTML]{63A8D3}} \color[HTML]{000000} 67.26 & {\cellcolor[HTML]{B4D3E9}} \color[HTML]{000000} 47.44 & {\cellcolor[HTML]{87BDDC}} \color[HTML]{000000} 58.97 & {\cellcolor[HTML]{65AAD4}} \color[HTML]{000000} 66.55 & {\cellcolor[HTML]{5DA5D1}} \color[HTML]{000000} 68.84 \\
Macro Precision & {\cellcolor[HTML]{71B1D7}} \color[HTML]{000000} 63.64 & {\cellcolor[HTML]{5AA2CF}} \color[HTML]{000000} 69.69 & {\cellcolor[HTML]{C7DCEF}} \color[HTML]{000000} 42.14 & {\cellcolor[HTML]{B4D3E9}} \color[HTML]{000000} 47.71 & {\cellcolor[HTML]{65AAD4}} \color[HTML]{000000} 66.59 & {\cellcolor[HTML]{529DCC}} \color[HTML]{000000} 71.97 \\
Macro Recall & {\cellcolor[HTML]{5DA5D1}} \color[HTML]{000000} 68.75 & {\cellcolor[HTML]{61A7D2}} \color[HTML]{000000} 67.78 & {\cellcolor[HTML]{8DC1DD}} \color[HTML]{000000} 57.60 & {\cellcolor[HTML]{9CC9E1}} \color[HTML]{000000} 54.29 & {\cellcolor[HTML]{57A0CE}} \color[HTML]{000000} 70.31 & {\cellcolor[HTML]{5BA3D0}} \color[HTML]{000000} 69.46 \\
Macro F1 & {\cellcolor[HTML]{6AAED6}} \color[HTML]{000000} 65.34 & {\cellcolor[HTML]{5DA5D1}} \color[HTML]{000000} 68.59 & {\cellcolor[HTML]{C2D9EE}} \color[HTML]{000000} 43.65 & {\cellcolor[HTML]{AFD1E7}} \color[HTML]{000000} 49.02 & {\cellcolor[HTML]{60A7D2}} \color[HTML]{000000} 67.82 & {\cellcolor[HTML]{57A0CE}} \color[HTML]{000000} 70.41 \\
Complicated F1 & {\cellcolor[HTML]{B0D2E7}} \color[HTML]{000000} 48.76 & {\cellcolor[HTML]{77B5D9}} \color[HTML]{000000} 62.32 & {\cellcolor[HTML]{A5CDE3}} \color[HTML]{000000} 51.91 & {\cellcolor[HTML]{5FA6D1}} \color[HTML]{000000} 68.28 & {\cellcolor[HTML]{B4D3E9}} \color[HTML]{000000} 47.73 & {\cellcolor[HTML]{7FB9DA}} \color[HTML]{000000} 60.46 \\
E F1 & {\cellcolor[HTML]{5AA2CF}} \color[HTML]{000000} 69.59 & {\cellcolor[HTML]{5DA5D1}} \color[HTML]{000000} 68.69 & {\cellcolor[HTML]{A9CFE5}} \color[HTML]{000000} 50.93 & {\cellcolor[HTML]{B2D2E8}} \color[HTML]{000000} 48.42 & {\cellcolor[HTML]{4E9ACB}} \color[HTML]{000000} 72.92 & {\cellcolor[HTML]{519CCC}} \color[HTML]{000000} 72.27 \\
N F1 & {\cellcolor[HTML]{68ACD5}} \color[HTML]{000000} 65.91 & {\cellcolor[HTML]{61A7D2}} \color[HTML]{000000} 67.50 & {\cellcolor[HTML]{D6E5F4}} \color[HTML]{000000} 35.43 & {\cellcolor[HTML]{C6DBEF}} \color[HTML]{000000} 42.77 & {\cellcolor[HTML]{5CA4D0}} \color[HTML]{000000} 69.20 & {\cellcolor[HTML]{5AA2CF}} \color[HTML]{000000} 69.67 \\
C F1 & {\cellcolor[HTML]{3F8FC5}} \color[HTML]{000000} 77.11 & {\cellcolor[HTML]{4493C7}} \color[HTML]{000000} 75.84 & {\cellcolor[HTML]{D3E4F3}} \color[HTML]{000000} 36.32 & {\cellcolor[HTML]{D3E3F3}} \color[HTML]{000000} 36.64 & {\cellcolor[HTML]{3383BE}} \color[HTML]{000000} 81.45 & {\cellcolor[HTML]{3989C1}} \color[HTML]{000000} 79.23 \\
\bottomrule
\end{tabular}
}
\caption{Left: 4-way classification performance on the test set. Right: Performance on the two subsets of the test set, Chaos and Original MNLI. Darker color indicates higher performance.}
\label{tab:performance}
\end{table}

\paragraph{Results}
Table~\ref{tab:performance} shows accuracy, macro F1, and F1 for each class.
Each score is the average from three random initializations. The macro F1 of our model is 68.59\% (vs.\ 65.34\% for MixUp), which is on par with previous work
\citep{zhang-de-marneffe-2021-identifying}, but with room for improvement. Our model generally outperforms the baseline, suggesting that
training with categorical labels is beneficial for predicting categorical labels.

\paragraph{``Complicated'' is most confused}
The model performs worse on the
Complicated label as opposed to the other three NLI labels.
This is consistent with \cite{kenyon-dean-etal-2018-sentiment}'s observation: the Complicated class is hard to model, due to its heterogeneity, as we saw in
Section~\ref{section:taxonomy}.
This is also shown in the confusion matrix in Table~\ref{tab:confusion_matrix}.
Conversely, there
are little errors among the three original NLI classes, which is partly due to the
stringent threshold we used to identify items on which we take the majority vote.

\paragraph{100 annotations are better}
Since the Complicated label is the most confused, we investigate where the confusion comes from. We partition the test set by whether the label comes from the ChaosNLI 100 annotations or the original MNLI 5 annotations, and compare the Complicated F1 on each subset.
Table~\ref{tab:performance} shows that for Complicated F1, the model performs much better on the Chaos subset with labels from 100 annotations than on the Original MNLI subset with labels from 5 annotations.
This suggests that 100 annotations provide clearer training signals and are more informative as to whether the items exhibit disagreement.

\begin{table}
\centering
\resizebox{\linewidth}{!}{
\begin{tabular}{p{1em}r|rrrr|r}
\toprule
& &\multicolumn{4}{c|}{Prediction} & \\
& & E & N & C & Complicated & All \\
\midrule
&E & {\cellcolor[HTML]{08306B}} \color[HTML]{F1F1F1} 274 & {\cellcolor[HTML]{F4F9FE}} \color[HTML]{000000} 10 & {\cellcolor[HTML]{F7FBFF}} \color[HTML]{000000} 3 & {\cellcolor[HTML]{EBF3FB}} \color[HTML]{000000} 108 & 395 \\
&N & {\cellcolor[HTML]{F7FBFF}} \color[HTML]{000000} 3 & {\cellcolor[HTML]{08306B}} \color[HTML]{F1F1F1} 257 & {\cellcolor[HTML]{F6FAFF}} \color[HTML]{000000} 5 & {\cellcolor[HTML]{E0ECF8}} \color[HTML]{000000} 130 & 395 \\
&C & {\cellcolor[HTML]{F3F8FE}} \color[HTML]{000000} 9 & {\cellcolor[HTML]{F7FBFF}} \color[HTML]{000000} 6 & {\cellcolor[HTML]{08306B}} \color[HTML]{F1F1F1} 297 & {\cellcolor[HTML]{F7FBFF}} \color[HTML]{000000} 83 & 395 \\
\parbox[t]{2mm}{\multirow{-4}{*}{\rotatebox[origin=c]{90}{Gold}}}
&Complicated & {\cellcolor[HTML]{8DC1DD}} \color[HTML]{000000} 116 & {\cellcolor[HTML]{AFD1E7}} \color[HTML]{000000} 87 & {\cellcolor[HTML]{C7DBEF}} \color[HTML]{000000} 76 & {\cellcolor[HTML]{08306B}} \color[HTML]{F1F1F1} 492 & 771 \\
&All & 402 & 360 & 381 & 813 & 1956 \\
\bottomrule
\end{tabular}
}
\caption{Confusion matrix of the 4-way classification predictions from the initialization with the highest macro F1. Darker color indicates higher numbers.}
\label{tab:confusion_matrix}
\end{table}

\begin{figure}
\includegraphics[width=\linewidth]{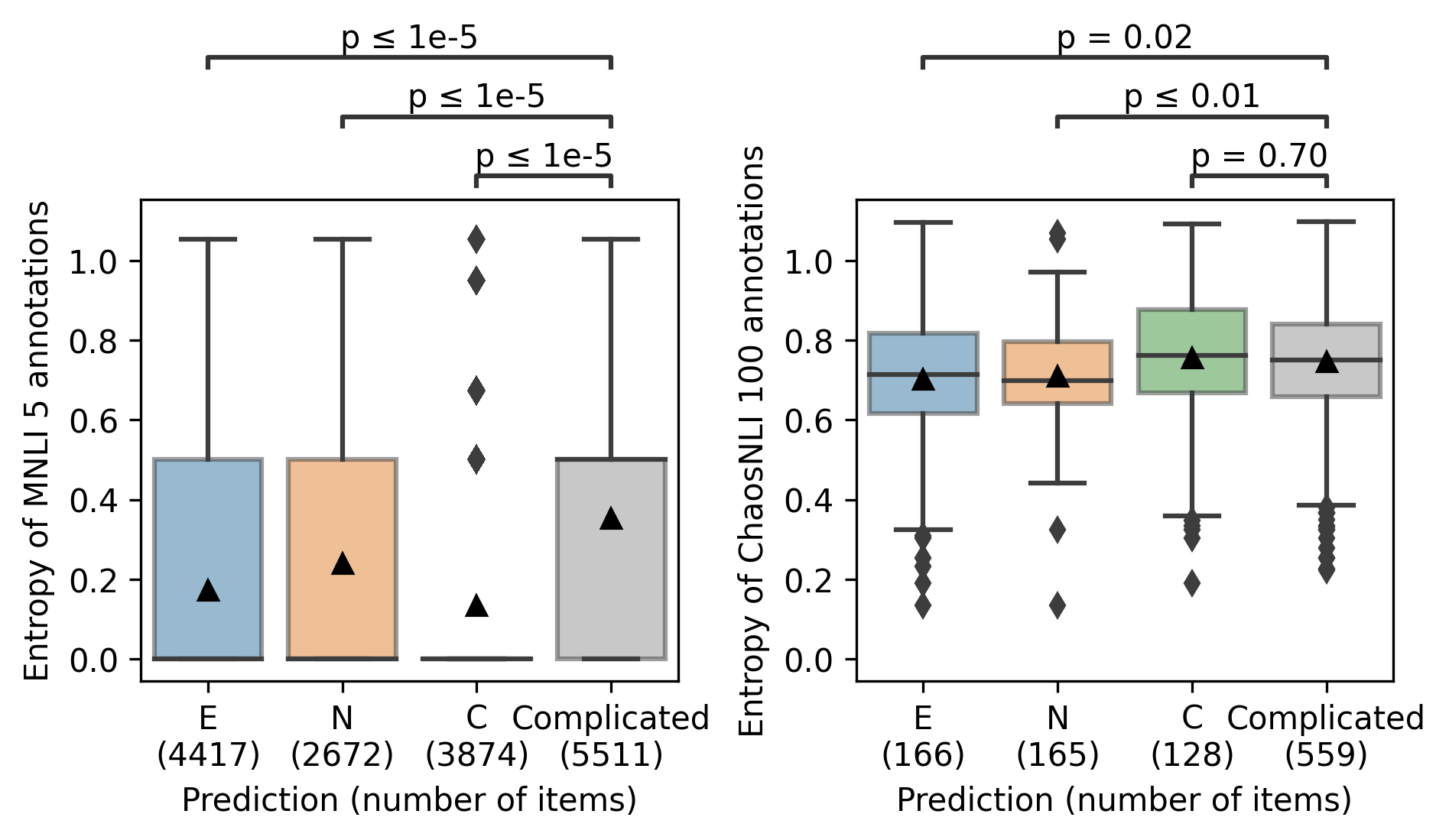}
\caption{Boxplots of annotation entropy (Left: from original MNLI 5
  annotations, Right: from ChaosNLI 100 annotations) by predicted label. Number of items shown in parentheses.
  Triangles indicate the means. P-values from Mann-Whitney two-sided test.
}
\label{fig:pred_entropy}
\end{figure}

\paragraph{Model predicts ``Complicated'' on high entropy items}
We can see how the model performs on a larger scale, not just limiting to items
where there is clear (dis)agreement.
For this analysis, we get the model predictions on the full MNLI matched and
mismatched dev sets, excluding the items used in our train and dev sets.
We compare the model predictions with the annotation entropy, shown in Figure~\ref{fig:pred_entropy}. Items predicted to be Complicated have significantly higher entropy than items predicted to be other labels (except for
predicted Contradiction and Complicated items from ChaosNLI).
This suggests that the model learned certain features associated with complicatedness.

\subsection{Multilabel classification}
\label{sec:multilabel_classification}
As mentioned in Section~\ref{sec:related}, the rationale for using a multilabel classification
approach is to get insight in the way in which an item is complicated.
Instead of choosing one of the three NLI labels (or four, including
Complicated), the model is to predict multiple of the three
NLI labels.

\paragraph{Model architecture}
To perform multilabel classification with 3 labels,
we make minimal changes to the standard method for fine-tuning RoBERTa for 3-way
classification.
We predict each E/N/C label independently, by applying the sigmoid function on top of the 3 logits given by the
MLP classifier on top of RoBERTa to obtain probabilities associated with each label.
We take the label to be present if its probability is greater than 0.5. The model is trained with a cross entropy loss.

\paragraph{Training procedure}
We used an initial learning rate of 5e-6, with learning rate decay by 0.8 times if dev F1 does not improve for one epoch.
We trained for up to 30 epochs, with early stopping used if dev F1 does not improve for 10 epochs.

\paragraph{Results}
Table~\ref{tab:multilabel_scores} gives the macro
precision, recall, and F1, and the exact match accuracy partitioned by the number of gold labels (1/2/3 Labels Accuracy) for the test set and for its subsets.
Our model has a higher F1 score than the baseline but a lower precision.
The baseline model is more successful at items on which annotators agree (higher 1
Label Accuracy), while our model performs better on items with disagreement (2/3
Labels Accuracy).

Comparing the two test set subsets, we see the same pattern as in the 4-way results: on disagreement items, our model performs better (higher 2/3 Labels Accuracy) on the Chaos subset than on the Orig subset. This corroborates the finding from the 4-way classification that
100 annotations give a better indication of complicatedness.

\paragraph{Multilabel model is more expressive}
The accuracy decreases from the 4-way classification setup, which is expected
since the number of possible labels increased from 4 to 7 (all possible
combinations of the 3 labels).
However, the macro recall increases compared to the 4-way classification (83.35 vs.\ 67.78),
possibly as a result of more expressivity in the model output and not
having one challenging and heterogeneous class.
We also see this more concretely
in the contingency table of the 4-way model vs.\ the multilabel model predictions (Table~\ref{tab:pred_contingency}): when the multilabel
model predicts more than one label, the 4-way model often predicts the Complicated class or one of the
labels predicted by the multilabel model.
In other words, the 4-way model may miss one or more labels while the
multilabel model can identify all of them.

\begin{table}
\centering
 \resizebox{\linewidth}{!}{
\begin{tabular}{lrr|rrrr}
\toprule
 & \multicolumn{2}{c|}{Chaos+Orig} & \multicolumn{2}{c}{Chaos} & \multicolumn{2}{c}{Orig} \\
 & MixUp & Our & MixUp & Our & MixUp & Our \\
\midrule
  Accuracy & {\cellcolor[HTML]{D2E3F3}} \color[HTML]{000000} 58.59 & {\cellcolor[HTML]{CEE0F2}} \color[HTML]{000000} 59.44 & {\cellcolor[HTML]{F7FBFF}} \color[HTML]{000000} 38.14 & {\cellcolor[HTML]{F7FBFF}} \color[HTML]{000000} 44.98 & {\cellcolor[HTML]{BED8EC}} \color[HTML]{000000} 62.47 & {\cellcolor[HTML]{BFD8ED}} \color[HTML]{000000} 62.19 \\
Macro Precision & {\cellcolor[HTML]{1F6EB3}} \color[HTML]{000000} 84.25 & {\cellcolor[HTML]{3080BD}} \color[HTML]{000000} 81.19 & {\cellcolor[HTML]{2777B8}} \color[HTML]{000000} 82.52 & {\cellcolor[HTML]{2C7CBA}} \color[HTML]{000000} 81.98 & {\cellcolor[HTML]{1D6CB1}} \color[HTML]{000000} 84.60 & {\cellcolor[HTML]{3181BD}} \color[HTML]{000000} 80.99 \\
Macro Recall & {\cellcolor[HTML]{4B98CA}} \color[HTML]{000000} 76.81 & {\cellcolor[HTML]{2373B6}} \color[HTML]{000000} 83.35 & {\cellcolor[HTML]{A0CBE2}} \color[HTML]{000000} 66.65 & {\cellcolor[HTML]{4896C8}} \color[HTML]{000000} 77.35 & {\cellcolor[HTML]{3A8AC2}} \color[HTML]{000000} 79.41 & {\cellcolor[HTML]{1C6AB0}} \color[HTML]{000000} 84.92 \\
Macro F1 & {\cellcolor[HTML]{3585BF}} \color[HTML]{000000} 80.29 & {\cellcolor[HTML]{2A7AB9}} \color[HTML]{000000} 82.17 & {\cellcolor[HTML]{63A8D3}} \color[HTML]{000000} 73.62 & {\cellcolor[HTML]{3A8AC2}} \color[HTML]{000000} 79.48 & {\cellcolor[HTML]{2D7DBB}} \color[HTML]{000000} 81.79 & {\cellcolor[HTML]{2676B8}} \color[HTML]{000000} 82.76 \\
1 Label Accuracy & {\cellcolor[HTML]{4493C7}} \color[HTML]{000000} 77.86 & {\cellcolor[HTML]{89BEDC}} \color[HTML]{000000} 69.23 & {\cellcolor[HTML]{D6E5F4}} \color[HTML]{000000} 57.66 & {\cellcolor[HTML]{F7FBFF}} \color[HTML]{000000} 45.05 & {\cellcolor[HTML]{3787C0}} \color[HTML]{000000} 79.95 & {\cellcolor[HTML]{72B2D8}} \color[HTML]{000000} 71.73 \\
2 Labels Accuracy & {\cellcolor[HTML]{F7FBFF}} \color[HTML]{000000} 30.20 & {\cellcolor[HTML]{F7FBFF}} \color[HTML]{000000} 46.16 & {\cellcolor[HTML]{F7FBFF}} \color[HTML]{000000} 32.52 & {\cellcolor[HTML]{EAF3FB}} \color[HTML]{000000} 52.97 & {\cellcolor[HTML]{F7FBFF}} \color[HTML]{000000} 29.53 & {\cellcolor[HTML]{F7FBFF}} \color[HTML]{000000} 44.21 \\
3 Labels Accuracy & {\cellcolor[HTML]{F7FBFF}} \color[HTML]{000000} 5.26 & {\cellcolor[HTML]{F7FBFF}} \color[HTML]{000000} 10.53 & {\cellcolor[HTML]{F7FBFF}} \color[HTML]{000000} 5.26 & {\cellcolor[HTML]{F7FBFF}} \color[HTML]{000000} 10.53 & {\cellcolor[HTML]{F7FBFF}} \color[HTML]{000000} 0.00 & {\cellcolor[HTML]{F7FBFF}} \color[HTML]{000000} 0.00 \\
\bottomrule
\end{tabular}
}
\caption{Left: Multilabel classification performance on the test set.
  Right: Performance on the two test set subsets. The Orig subset does not have any items with all three labels present.}
\label{tab:multilabel_scores}
\end{table}

\begin{table}
\resizebox{\linewidth}{!}{
\begin{tabular}{lrrrrrrr}
\toprule
 &     E &     N &     C &    EN &    NC &   EC &  ENC \\
\midrule
E           &  3718 &     8 &    19 &   503 &     5 &  178 &   32 \\
N           &     9 &  2076 &     4 &   467 &   275 &    0 &    8 \\
C           &     7 &     6 &  3170 &    35 &   479 &  135 &   49 \\
Complicated &   395 &   838 &   300 &  1995 &  1289 &  162 &  312 \\
\bottomrule
\end{tabular}
}
\caption{Contingency matrix of the 4-way classification vs.\ the
  multilabel predictions, on the full MNLI dev sets (excluding items used in
  our train/dev sets).}
\label{tab:pred_contingency}
\end{table}

\newcounter{errorexample}
\renewcommand{\theerrorexample}{\arabic{errorexample}}
\newcommand{\errorex}{\leavevmode\refstepcounter{errorexample} \theerrorexample}

\begin{table*}
  \small
  \resizebox{\textwidth}{!}{
\begin{tabular}[]{@{}p{.3cm}p{10cm}p{6.7cm}p{1.6cm}p{2.5cm}}
    \toprule
&Premise                                                                                                                           & Hypothesis                                            & [E,N,C]      & Predictions \\
    \midrule
&  \textbf{Probabilistic Enrichment} \\
  \midrule
\errorex \label{ex:er_prob_church}&Oh, sorry, wrong church. & He or she entered the wrong church. & [82, 17, 1]	& Complicated / EN\\
\errorex \label{ex:er_prob}&There should be someone here who knew more of what was going on in this world than he did now.& He knew things, but hoped someone else knew more.&[82, 18, 0]	 & Complicated /	EN	 \\
\errorex&What am I to do with them afterwards?" & It is the narrator's responsibility to take care of them. & [15, 73, 12] & 	N / NC \\
\errorex&But they persevered, she said, firm and optimistic in their search, until they were finally allowed by a packed restaurant to eat their dinner off the floor.&	Because all of the seats were stolen, they had to eat off the floor.&	[23, 57, 20]	& Complicated / NC \\
  \midrule
&  \textbf{Coreference} \\
  \midrule
\errorex \label{ex:er_coref_wax}&The original wax models of the river gods are on display in the Civic Museum.                                                     & They have models made out of clay.                    & [5, 38, 57]  & C / C           \\
\errorex \label{ex:er_coref_sandoro}&Indeed, said San'doro.                                                                                                            & Indeed, they said.                                    & [52, 22, 26] & E / EN           \\
\errorex \label{ex:er_coref_jahangir}&This was built 15 years earlier by Jahangir's wife, Nur Jahan, for her father, who served as Mughal Prime Minister.	& Nur Jahan's husband Jahangir served as Mughal Prime Minister.	& [17, 17, 66] & Complicated / E \\
\errorex \label{ex:er_coref_cruise}&Cruises are available from the Bhansi Ghat, which is near the CityPalace.	&You can take cruises from Phoenix Arizona.&	[0, 51, 49]	&Complicated /NC	\\
  \midrule
&  \textbf{Accommodating Minimally Added Content} \\
  \midrule
\errorex\label{ex:er_accommodate_hilary}&The key question may be not what Hillary knew but when she knew it. &	According to
current reports, the question is not if, but when did Hillary know about it. &
[90, 9, 1]	& E / EN \\
\hspace{-1em}\errorex\label{ex:er_accommdoate_kal}&Four or five from the town rode past, routed by their diminished numbers and the fury of the Kal and Thorn. &	Kal and Thorn were furious at the villagers. &	[50, 41, 9]	& N / EN \\
    \bottomrule
\end{tabular}}
\caption{Examples from the categorization with ChaosNLI
  annotations and 4-way/multi-label model predictions.}
\label{tab:error}
\end{table*}

\paragraph{Takeaways}
Comparing with the MixUp baseline which is trained with soft-labels,
we see that training with categorical labels performs better in predicting
categorical labels.
Therefore, for downstream tasks where categorical information is needed,
training with categorical labels is recommended.
The multilabel model is more expressive, and
as we will show in Section~\ref{sec:error}, it provides fine-grained information that gives a better understanding of what the model has learned.
Our results suggest that the multilabel approach could potentially be used as
intrinsic evaluation for how well the model captures the judgments of the population.

\section{Error Analysis}
\label{sec:error}
\begin{figure}[t]
  \centering
  \includegraphics[width=\linewidth]{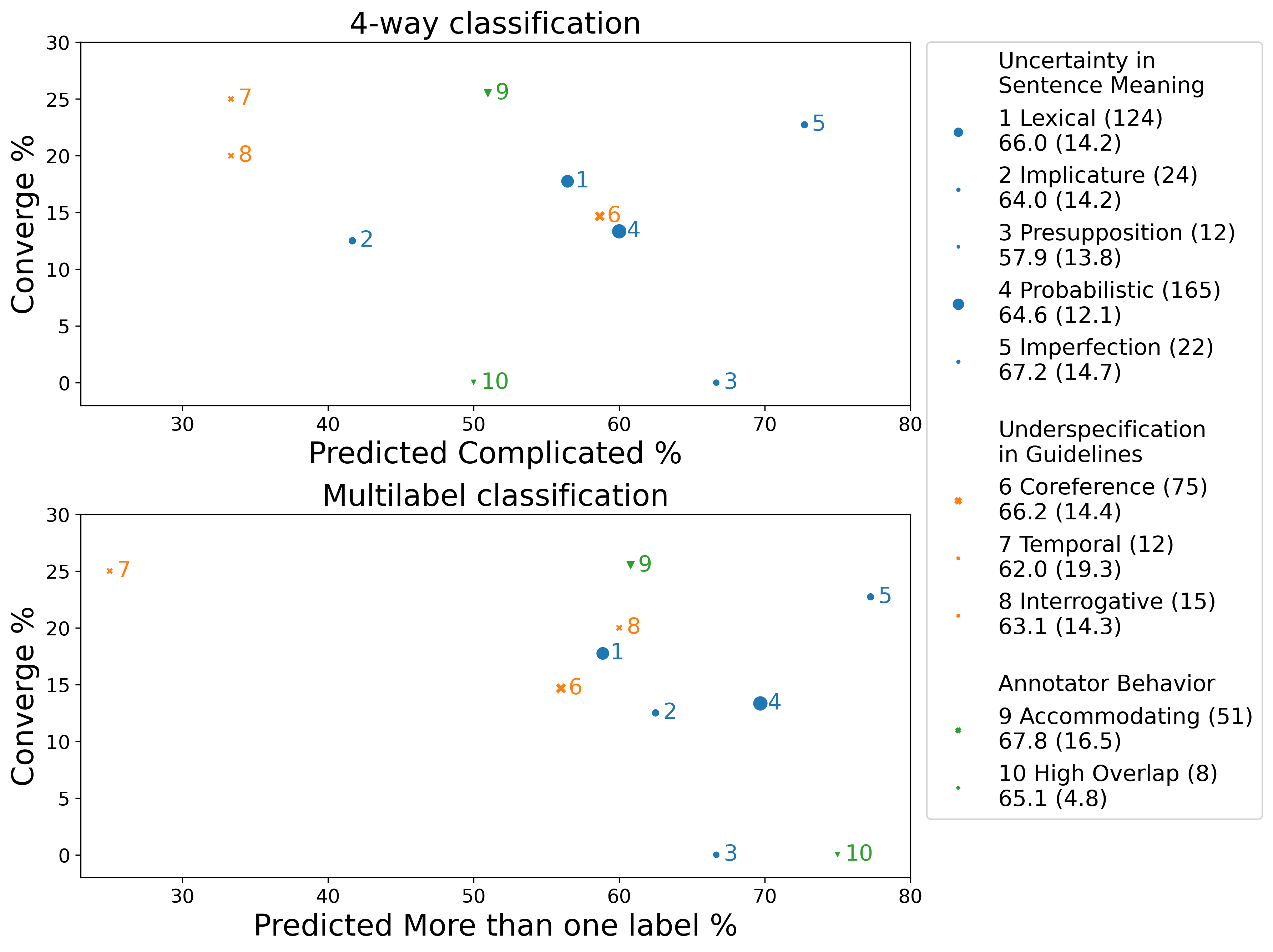}
  \caption{For each disagreement category, percentage of ChaosNLI items annotated with that category (number in parentheses) and having converging
   NLI annotations (>80 majority vote) vs.\ percentage
    predicted as Complicated in the 4-way setup or as having more than one label in the multilabel
    setup.
    Legend also gives mean majority vote in each category, with standard deviation in parentheses.
    }
  \label{fig:scatter}
\end{figure}

We analyze the model behavior with respect to the categories of disagreement
sources.
For each category, Figure~\ref{fig:scatter} gives the percentages of ChaosNLI items annotated with at least that category and having converging NLI interpretations (>80
agree on the NLI label) vs.\ percentages of items predicted to exhibit disagreement (Complicated by the 4-way model or
as having more than one label by the multilabel model).
Overall, a category with more agreement (higher majority vote) in the annotations tend to have less items predicted as exhibiting
disagreement.
This is expected given that
an item with convergence corresponds to not having disagreement as gold label, and the model performs well overall.

Comparing the two models, we see that all categories, except [7] Temporal Reference, are farther to the right in the Multilabel classification (bottom panel) whereas they are more
spread out in the 4-way classification (top panel), meaning that
the 4-way model predicts an agreement label (E/N/C) more often than the Multilabel model. 
This suggests that the 4-way model is more strongly tied to the convergence statistics and failing to detect potentials of disagreement.
It also aligns with the previous finding that the Multilabel model has higher recall of the range of interpretations.

Items in [3] Presupposition, [4] Probabilistic Enrichment, [5] Imperfection are often predicted in both setups to exhibit disagreement (they are to the right of both plots in Figure~\ref{fig:scatter}). [6] Coreference, [2] Implicature and [10] High Overlap also appear to the right, depending on the setup.
Among those categories, [3] Presupposition, [2] Implicature, [5] Imperfection and [10] High Overlap are associated with surface patterns, potentially making it easier for the models to learn that they often exhibit disagreement.
We thus take a closer look at the following categories, across all items annotated:
Probabilistic Enrichment, Coreference, and Accommodating Minimally Added Content (discussed in Section~\ref{findings}).

\paragraph{Probabilistic Enrichment}
The multilabel model predicts
68\% of the items annotated with Probabilistic Enrichment to have more than one NLI label. 
In particular, 36\% are predicted as EN and 27\% as
NC, corresponding the common patterns in Probabilistic Enrichment where the
enriched (not explicitly stated) inference leads to Entailment/Contradiction, and the Neutral label is warranted without enrichment.
We found that the multilabel model often predicts labels when they are only slightly below the threshold of 20 that we used to count a
label as present (items \ref{ex:er_prob_church} and \ref{ex:er_prob} in Table~\ref{tab:error}).
Even though in those cases the model is ``incorrect'' when calculating the metrics,
it shows that the model can retrieve subtle inferences.
In item \ref{ex:er_prob_church}, 17 annotators chose Neutral, while 82
chose Entailment: the premise does not mention \textit{entering} a church, but most annotators take that situation to be likely. The multilabel model is however predicting both Entailment and Neutral, accounting for the possible interpretations.

\paragraph{Coreference}

For items annotated with Coreference,
both models predict Entailment/Contradiction when the premise and hypothesis
share the same argument structure or involve simple word substitutions (e.g.\ \textit{wax/clay} in item \ref{ex:er_coref_wax} and \textit{San'doro/they} in item \ref{ex:er_coref_sandoro}, Table~\ref{tab:error}), which are features of unanimous Entailment/Contradiction.
This suggests that such predictions are influenced by the unanimous items.
The 4-way model tends to predict Complicated when items annotated with Coreference do not share any structure (as in items
\ref{ex:er_coref_jahangir} and \ref{ex:er_coref_cruise}).

\paragraph{Accommodating Minimally Added Content}
The multilabel model predicts 44\% of the items involving minimally
added content to have both Entailment and Neutral labels, and 76\% of the items to
have at least the Neutral label.
This is consistent with the majority of these items showing disagreement over Entailment and Neutral, and the sentences themselves exhibiting
features of Neutral (added content) and surface features of Entailment (high lexical overlap), as in items \ref{ex:er_accommodate_hilary} and \ref{ex:er_accommdoate_kal}.
In item \ref{ex:er_accommodate_hilary},
the multilabel model recovers a Neutral inference (the premise does not mention current reports), even when only 9 annotators chose the Neutral label.
This further illustrates that the multilabel model is better at recalling possible interpretations.

\section{Conclusion}
We examined why disagreement in NLI annotations occurs
and found that it arises out of all three components
of the annotation process.
We experimented with modeling NLI disagreement as 4-way and multilabel classifications, and showed that the multilabel model gives a better
recall of the range of interpretations.
We hope our findings will shed light on how to improve the NLI annotation process, e.g.\ ways to specify the guidelines to reduce
disagreement or introduce contexts that resolve underspecification, ways to gather enough annotations to cover the possible interpretations, as well as
ways to model NLI without the single ground truth assumption.

\section*{Acknowledgement}
We thank TACL editorial assistant Cindy Robinson and action editor Anette Frank for the time they committed to the review process. We thank Anette Frank for her clear and detailed editor letter, as well as the anonymous reviewers for their insightful feedback. We also thank Micha Elsner gand Michael White, and members of the OSU Pragmatics and Clippers discussion groups for their suggestions and comments, and especially Angélica Aviles Bosques for her help with the annotation. This material is based upon work supported by the National Science Foundation under grant no.\ IIS-1845122. Marie-Catherine de Marneffe is a Research Associate of the Fonds de la Recherche Scientifique -- FNRS.

\bibliography{anthology,custom,Disagreement}
\bibliographystyle{acl_natbib}

\end{document}